\documentclass[conference]{IEEEtran}
\usepackage[pdftex]{graphicx}
\usepackage{epstopdf}
\usepackage[cmex10]{amsmath}
\usepackage{cases,amssymb}
\usepackage{subeqnarray,subfigure}
\usepackage[linesnumbered,ruled,vlined]{algorithm2e}
\usepackage{cite,setspace,bm,enumerate}
\usepackage{amsmath,color}
\usepackage{amsfonts,caption}
\usepackage{bbm}

\DeclareMathOperator*{\argmax}{arg\,max}

\begin{document}

\title{Edge Computing Enabled Real-Time Video Analysis via Adaptive Spatial-Temporal Semantic Filtering}

\author{ 
	\small Xiang Chen$^{\ast}$, Wenjie Zhu$^{\ast}$, Jiayuan Chen$^{\ast}$, Tong Zhang$^{\ast}$, Changyan Yi$^{\ast}$ and Jun Cai$^{\dagger}$\\
	\IEEEauthorblockA{\text{\small $^{\ast}$College of Computer Science and Technology, Nanjing University of Aeronautics and Astronautics, Nanjing, China} \\
		\small $^{\dagger}$Department of Electrical and Computer Engineering, Concordia University, Montr\'{e}al, QC, H3G 1M8, Canada \\
		\text{\small Email: \{chenxiang, wenjie.zhu, jiayuan.chen, zhangt, changyan.yi\}@nuaa.edu.cn,  jun.cai@concordia.ca}\\
	}
}

\IEEEtitleabstractindextext{%

%novel double deep Q network-contextual multi-armed bandit (DDQN-CMAB) reinforcement learning (DCRL) framework

%\vspace{-.5em}

\begin{abstract}
This paper proposes a novel edge computing enabled real-time video analysis system for intelligent visual devices. The proposed system consists of a tracking-assisted object detection module (TAODM) and a region of interesting module (ROIM). TAODM adaptively determines the offloading decision to process each video frame locally with a tracking algorithm or to offload it to the edge server inferred by an object detection model. ROIM determines each offloading frame’s resolution and detection model configuration to ensure that the analysis results can return in time. TAODM and ROIM interact jointly to filter the repetitive spatial-temporal semantic information to maximize the processing rate while ensuring high video analysis accuracy. Unlike most existing works, this paper investigates the real-time video analysis systems where the intelligent visual device connects to the edge server through a wireless network with fluctuating network conditions. We decompose the real-time video analysis problem into the offloading decision and configurations selection sub-problems. To solve these two sub-problems, we introduce a double deep Q network (DDQN) based offloading approach and a contextual multi-armed bandit (CMAB) based adaptive configurations selection approach, respectively. A DDQN-CMAB reinforcement learning (DCRL) training framework is further developed to integrate these two approaches to improve the overall video analyzing performance. Extensive simulations are conducted to evaluate the performance of the proposed solution, and demonstrate its superiority over counterparts.
\end{abstract}

%\vspace{-2em}

%\begin{IEEEkeywords}
%Dynamic application placement, vacation queue, edge computing, latency minimization.
%\end{IEEEkeywords}
}

\maketitle
\IEEEdisplaynontitleabstractindextext
\IEEEpeerreviewmaketitle

\section{Introduction}\label{introduction}

\IEEEPARstart{W}{ith} the fast burgeoning of the Internet of Things (IoT), intelligent visual devices (e.g., smartphones, auto-driving vehicle, unmanned aerial vehicle) have encouraged the emergence of far-reaching innovative mobile applications, ranging from autonomous systems to extended reality\cite{10238695}. These intelligent visual devices capture surroundings and generate skyrocketing amounts of video data that need to be processed in real-time. For example, self-driving applications leverage intelligent camera to accurately detect the lanes, cars, and pedestrians to avoid collisions. However, processing such massive computation-intensive and delay-sensitive video analysis tasks locally is significantly challenging due to the limited resources of intelligent visual devices\cite{dlvat}. Although an alternative way is to offload tasks to powerful cloud servers, it will bring unacceptable delays because they are typically deployed far away from the devices. 
%Thus, offloading computing tasks to the powerful server at the network edge emerges. However, Due to the severe delay and accuracy requirements of real-time video analysis tasks, this problem is still very challenging but common network conditions (e.g., cellular networks), even with edge computing\cite{wave}.

%Intuitively, processing all video frames on the device side in real time is best. Unfortunately, it is impossible on a weak resources device for the advanced deep neural network (DNN) models need large RAM and computing power\cite{splitplace}. Meanwhile, it is tough to offload all video frames to an edge server in real time due to the limited uplink bandwidth of cellular networks and the vast amount of data. Furthermore, the cellular network condition is very volatile and fluctuating; the coefficient of variation for LTE is up to 36\%, while WiFi is 2.4GHz and 1\% for 5GHz\cite{wave}. Besides, the cellular network expense of users is based on the data traffic they use, and the computing resources of edge servers are not accessible. Fortunately, the vast repetitive spatial-temporal semantic information between video frames gives us a glimmer of hope\cite{respire}.
% 语义过滤没有切入进来
Fortunately, the paradigm of edge computing\cite{10173745}, which deploys powerful edge servers at the edge of networks, provides pervasive, reliable and fast-responsive computing services for the devices, and thereby offloading video analysis tasks to edge servers is a promising way. However, due to the limited uplink bandwidth and volatility of the celluar network, it is difficult to offload all video frames with massive sizes to edge servers in real time\cite{wave}. Besides, vast repetitive spatial-temporal semantic information among video frames, which generates large transmission delays, severely impacts real-time analyzing performance. Moreover, due to the huge amount of video traffic, users have to pay huge expense on data traffic. Above factors motivate us to design an adaptive spatial-temporal semantic filtering based video analysis offloading strategy with low bandwidth cost and high detection accuracy under the volatile network conditions.

Nevertheless, designing such a strategy is significantly challenging due to the following reasons: i) in particular, real-time video will generate massive traffic per unit of time, resulting in the visual device being unable to transmit all the images to the edge server (ES) timely, and ii) constantly growing intelligent visual devices with frequent video analysis task requests would drain the resources of edge servers fast. In addition, there are considerable differences in computing power among various intelligent visual devices, leading to the performance of the sames algorithm varying wildly. 

Aiming to solve the above problems, we study a long-term semantic-filtering offloading optimization problem to maximize the processing rate, i.e., the number of frames of detection results returned in time,  and the detection accuracy under the time-varying network conditions. It is observed that this problem involves two coupled intricate sub-problems, an offloading decision problem and a configurations selection problem. On this basis, we propose a double deep Q network (DDQN) based offloading decision scheme and a contextual multi-armed bandit (CMAB) based adaptive configurations selection scheme for these two sub-problems, respectively. To obtain the optimal solution, we further propose a two-layer training framework, namely DDQN-CMAB reinforcement learning (DCRL) framework, for jointly solving these two sub-problems effectively. The main contributions are summarized as follows.
\begin{itemize}
	\item A novel edge computing enabled real-time video analysis system based on spatial-temporal semantic filtering is proposed, which realizes real-time object detection with high accuracy under volatile and fluctuating network conditions.
	\item A DCRL framework is designed for solving the overall semantic-filtering offloading optimization problem under unpredictable network conditions.
	\item Simulations are conducted to examine the performance of the proposed DCRL framework on the multi-camera pedestrian video dataset\cite{crossdataset}.
\end{itemize}

The rest of this paper is organized as follows: Section \ref{system_model_and_problem_formulation} presents the system model and problem formulation. In Section \ref{vacation_queue_based_performance_analysis}, two specific approaches are proposed to solve the decomposed sub-problems, and a DCRL framework is
proposed to integrate these two approaches. Simulation results are given in Section \ref{simulation_results}, followed by conclusions in Section \ref{conclusion}.

\section{System Model and Problem Formulation}\label{system_model_and_problem_formulation}

\subsection{System Model}\label{overview_of_the_system}

% 图太小气，和标题呼应，画好看一些
\begin{figure}[!t]
	\centering
	\includegraphics[width=3.5in]{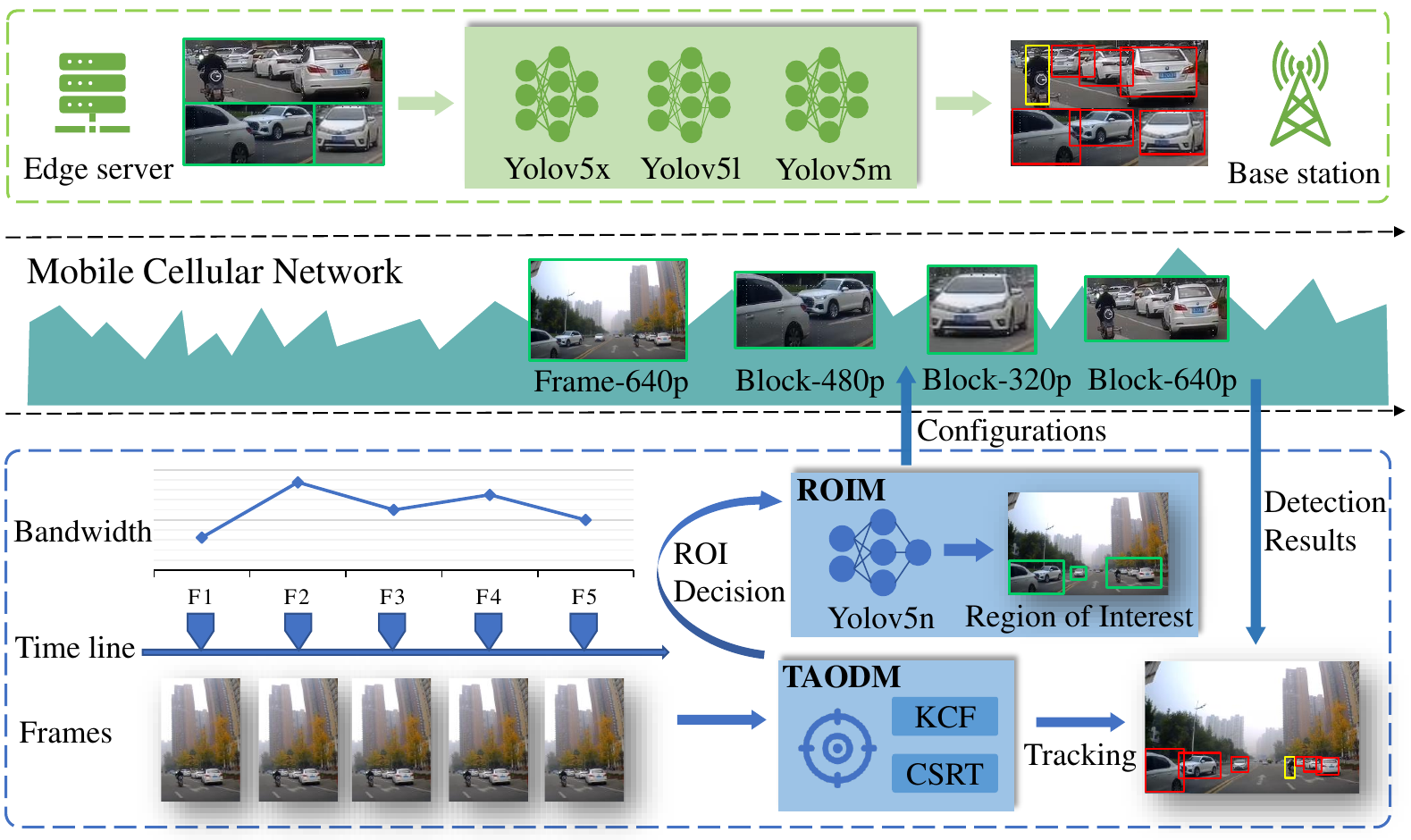}\\
	\caption{Overview of proposed real-time video analysis system.}
	\label{model}
\end{figure}

Consider an collaborative edge-device real-time video analysis system, as illustrated in Fig. \ref{model}, which consists of a tracking-assisted object detection module (TAODM) and a region of interesting module (ROIM) deployed on a mobile intelligent visual device to process road condition video stream and a dedicated ES with various advanced object detection models deployed at the base station. Besides, a frame-based time-slotted operation framework is studied to depict time-varying uncertainties of the video analysis system, where $t \in \{1, 2, \dots, T\}$ denotes the index of time slots and each time slot represents a decision interval. In the proposed edge computing enabled real-time video analysis system, an intelligent visual device with computing power $f^{device}$ connects to an ES with computing power $f^{edge}$ through the cellular network with bandwidth $b_t$. The intelligent visual device captures the real-time video stream and generates a real-time frame in each time slot by sampling. The real-time frame is denoted as $f_t$, which is assumed to be generated at beginning of time slot $t$ and will be processed by TAODM. TAODM adaptively determines the offloading mode to process the frame locally with a tracking algorithm or to offload it to the ES. If the latter, the offloaded frame would be further processed by ROIM, which determines its resolution and detection model configuration according to the information density and bandwidth for ensuring the analysis results return in time. Since the tracking algorithm requires the detection results from ES to initialize tracking targets, TAODM and ROIM are highly coupled. TAODM and ROIM interact jointly to maximize the processing rate while ensuring high video analysis accuracy. 
%Finally, real-time frame analyzed results are acquired for the device to assist in auto-driving or rescuing in dangerous environments.

%In the TAODM process for frame $f_t$, we use a set $\mathcal{K} = \{Skip, KCF, CSRT\}$ to denote various tracking modes. KCF\cite{kcf} and CSRT\cite{csrt} are the best known trackers and work fast on weak computing devices as well. 
Specifically, three decisions would be made by TAODM, namely,  i) offloading decision $\alpha_t \in \{0, 1\}$, which indicates the frame $f_t$ to be processed locally or to be offloaded to the ES ii) ROI decision $\beta_t \in \{0, 1\}$ indicates whether to adopt ROI extraction or not and iii) the tracking mode $k \in \mathcal{K}$ to be adopted, where $\mathcal{K} = \{\emph{Skip, KCF, CSRT}\}$, and the \emph{Skip} mode would directly reuse the detection result of the last frame if the current frame has tiny new semantic information compared to the last one, and \emph{KCF}\cite{kcf} and \emph{CSRT}\cite{csrt} are well-known trackers working well in resource-limited IoT devices.
When $\alpha_t=0$, the frame will be processed by TAODM with tracking mode $k \in \mathcal{K}$, and we use $acc(k)$ to represent the processing accuracy of tracking mode $k$. Additionally, the processing time of tracking mode $k$ can be calculated as $l_{t}^{pro,loc} = u^{k}d_t /f^{device}$, where $u^{k}$ indicates the computing intensity of tracking mode $k$ and $d_t$ represents the data size of frame $f_t$. Note that, $l^{pro,loc} = 0$ when $k = skip$. When $\alpha_t=1$, the ROI extraction decision $\beta_t \in \{0, 1\}$ will be made according to the during time slot $t$  bandwidth and then the frame $f_t$ is further processed by ROIM.  
 %and a local lightweight deep neural network (DNN) model detect the frame with inference delay $l_t^{dnn,loc}$.

In ROIM, the frame $f_t$ would firstly go through the process of rough pre-detection by a local lightweight deep neural network (DNN) for obtaining the bounding boxes and the amount of them, which indicate the information density $n_{t}$, with inference time $l_t^{dnn, loc}$. Then, if $\beta_t =1$, all bounding boxes will be merged into several ROI blocks with extraction time $l_{t}^{roi,loc} = {u^{roi} d_t}/{f^{device}}$, where $u^{roi}$ represents the computing intensity of ROI . After that, only those ROI blocks are transmitted to the edge server for further processing. When $\beta_t = 0$, the whole frame will be transmitted instead. Considering the inference requests of a frame can be served by different detection models, we use the set $\mathcal{M} = \{Yolov5x, Yolov5l, Yolov5m\}$ to denote the detection models held on the ES. YOLOv5 is the most popular object detection algorithm nowadays and provides various size models to meet different scenarios. Besides, we use a set $\mathcal{R} = \{640p, 480p, 320p\}$ to denote the set of offloading resolutions to accelerate the model inference and reduce the transmission delay.
Thus, ROIM would make two decisions according to the information density $n_{t,i}$ and bandwidth $b_t$ before transmitting the frame to the edge servers for real-time performance, i.e., i) the model $m \in \mathcal{M}$ selected for object detection in the edge server and ii) offloading resolution $r \in \mathcal{R}$ for each frame or block.
Then, based on the decision of offloading resolution $r \in \mathcal{R}$, frame or all blocks are resized to the offloading resolution and the size of offloading data can be obtained as $d_t = \sum_{i = 1}^{s_t}{\tau b(r_{t,i})}$, where $s_t$ represents the number of blocks, $\tau$ is the data size of one pixel and $b(r_{t,i})$ is the pixel number of block $i$ in resolution $r$, meanwhile, transmitting a whole frame is treated as transmitting a block for simplification. Finally, the frame or all ROI blocks with the corresponding model configurations has the transmission delay $l_{t}^{trans} ={d_t}/{b_t}$, and detected by the advanced detection model. 
% Finally, real-time frame analyzed results are acquired for the device to assist in auto-driving or rescuing in dangerous environments.

Based on above formulation, the completion time of frame $f_t$ can be expressed as 
\begin{align}
	l^{total}_t = & \alpha_t(\beta_t l_{t}^{roi,loc} + l_t^{dnn,loc} + l_{t}^{trans} + l_{t}^{pro,edge}) \notag\\
		&+ (1-\alpha_t)l_{t}^{pro,loc},
\end{align}
where $l_{t}^{pro,edge} = {u^{m,r} d_t}/{f^{edge}}$ is the inference delay in the ES and $u^{m,r}$ represents the computing intensity of model $m$ in resolution $r$. The size of inference results is commonly negligible compared to that of the offloading frame, so we ignore the delay of inference results transmitting back to the intelligent visual device from the ES. Since the video frames need to be analyzed in real-time, each frame should be processed within a time duration $l^{max}$, which can be formulated as
\begin{equation}
	l^{total}_t \leq l^{max},
\end{equation}
where $l^{max}$ is the length of a time slot.
Besides, the detection accuracy of video analysis system $acc_t$ can be mathematically expressed as
\begin{align}
	acc_t = & \frac{\alpha_t}{s_t} \sum\nolimits_{i=1}^{s_t} \sum\nolimits_{m \in \mathcal{M}} \sum\nolimits_{r\in \mathcal {R}} y_{t,i}^m z^{r}_{t,i} acc(m,r)\notag \\
		& + (1-\alpha_t) \sum\nolimits_{k \in \mathcal{K}}{x_t^kacc(k)},
\end{align}
where $s_t$ represents the number of blocks, $x_t^k \in \{0, 1\}$ indicates whether the tracking mode $k$ is selected ($x_t^k$ = 1 signifies the mode $k$ is selected, otherwise $x_t^k$ = 0). Similarly, $y_{t,i}^m \in \{0, 1\}$ and $z_{t,i}^r \in \{0, 1\}$ indicate whether the detection model configuration $m$ and offloading resolution $r$ are selected. 

Note that, different tracking modes have different processing time. \emph{Skip} mode has almost no latency, but detection accuracy degrades quickly when semantic information changes. The \emph{KCF} mode is faster with lower accuracy than the \emph{CSRT} mode. In addition, different configurations of detection models and offloading resolutions have different precision and inference delays. Adopting the advanced object detection model and high offloading resolution will have a better accuracy but lead to a longer inference delay and transmission delay, which damages the real-time performance. Meanwhile, selecting the entire frame can improve accuracy for that ROI extraction may miss some key information but also incurs a higher transmission delay, decreasing the real-time video analysis performance.

\subsection{Problem Formulation}\label{problem_formulation}

Since the quality of the proposed video analysis system highly depends on the frame processing rate (i.e., the frame number of detection results returned in time) and detection accuracy, in this work, we aim to maximize the system utility, which is a weighted sum of these two factors. Thus, the optimization problem of video analysis can be mathematically expressed as
\begin{align}
	&\max_{\{\alpha_t,\beta_{t},x_t^k,y_{t,i}^m,z_{t,i}^r\}}~ \frac {1}{T}\sum\nolimits_{t=1}^{T} q_{t} + \eta \frac {\sum\nolimits _{t=1}^{T}(q_t acc_t)}{\sum\nolimits_{t=1}^{T} q_{t}},\\
	&\quad \text {s.t.}~ \alpha_t, \beta_t \in \{0,1\},\\
	&\qquad ~(l^{total}_t - l^{max})q_t \leq 0, ~q_t \in \{0,1\},\\
	&\qquad ~\sum\nolimits_{k \in \mathcal{K}} x^{k}_{t}\leq 1, x^{k}_{t}\in \{0,1\},  ~k \in \mathcal{K},\\
	&\qquad ~\sum\nolimits_{m \in \mathcal{M}} y^{m}_{t,i}\leq 1,  y^{m}_{t,i}\in \{0,1\}, ~m \in \mathcal{M},\\
	&\qquad ~\sum\nolimits_{r\in \mathcal {R}} z^{r}_{t,i}\leq 1, z^{r}_{t,i}\in \{0,1\}, ~r \in \mathcal{R},\\
	&\qquad ~\sum\nolimits_{k \in \mathcal{K}} x^{k}_{t} = \alpha_t, ~k \in \mathcal{K},\\
	&\qquad ~\sum\nolimits_{k \in \mathcal{K}} x^{k}_{t} + \sum\nolimits_{m \in \mathcal{M}} y^{m}_{t,i} = 1, ~k \in \mathcal{K}, m \in \mathcal{M},\\
	&\qquad ~\sum\nolimits_{k \in \mathcal{K}} x^{k}_{t} + \sum\nolimits_{r\in \mathcal {R}} z^{r}_{t,i} = 1, ~k \in \mathcal{K}, r \in \mathcal{R},
\end{align}
where parameter $\eta >$ 0 is used to depict the weight of detection accuracy, and $q_t \in \{0, 1\}$ denotes whether frame $f_t$ is processed successfully in the time slot $t$. Specifically, if $l_t^{total} > l^{max}$, it means that frame $f_t$ is not processed completely within $l^{max}$ and will be abandoned, , and $q_t=0$ must be established to make constraint (6) true. If $l_t^{total} \leq l^{max}$, $q_t = 0$ means that the detection accuracy of frame $f_t$ is lower than a threshold thus is not successfully processed, while $q_t = 1$ means that frame $f_t$ is really successfully processed.
%$q_t = 0$ means that the frame $f_t$ is not processed completely within $l^{max}$ and will be abandoned, while $q_t = 1$ indicates whether to wait for the analysis result can decide by the video analysis system within $l_{max}$.Constraint (6) specifies that only the analysis results of a frame obtained within $l^{max}$ are counted as being successfully processed; 
Constraint (10) means a tracking mode is adopted when $\alpha_t = 1$. Otherwise, the configurations of models and resolutions will be chosen for offloading, represented by Constraint (11) and Constraint (12).  

Observe that the formulated optimization problem is a mixed-integer nonlinear programming problem, which is NP-hard problem \cite{edgeadaptor}. To address this problem, we partition the problem into two sub-problems, i.e., deciding the optimal offloading strategy for the frame, and selecting object detection model and offloading resolution configuration for each block. In the next section, a real-time DDQN-CMAB reinforcement learning framework is adopted to connect a DDQN-based offloading approach and a CMAB-based configuration selection approach to generate the optimal strategy for real-time video analysis.

\section{DDQN-CMAB Reinforcement Learning Strategy}\label{vacation_queue_based_performance_analysis}

\begin{algorithm}[!t]
	\small
	\caption{DDQN-based Offloading Algorithm} \label{algorithm1}
	\KwOut{The network parameters $\theta^*$}
	Initialize: Randomly initialize the parameters $\theta$; initialize the target Q-network parameter $\theta^\prime = \theta$; reset the replay buffer $B^d$ of experience playback; maximum training steps $LOOP$ and $loop = 0$.\\
	\For{$loop$ $\leq$ $LOOP$}{
		Obtain initial observation state $s_0$ and replay buffer $B^d$;\\
		\For{$t=1$ to $T$}{
			Observe the state $s_t$;\\
			\If{random $\geq$ $\epsilon^{ddqn}$}{
				$a_t \leftarrow \argmax_{a \in A_t}{Q(s_t, a_t; \theta_t)}$;
			} \Else{
				Randomly select $a_t$ from $A_t$;
			}
			Execute $a_t$ and receive reward $R$;\\
			Store tuple $(s_t, a_t, R)$ into $B^d$; \\
			Sample random batch of tuples from $B^d$; \\
			Compute the value of Q-traget;\\
			Update current Q-network parameters $\theta_t$ according to loss function;\\
			\If{$t \text{ mod } C = 0$}{
				Update target Q-network parameters $\theta^{\prime}_t \leftarrow \theta_t$;\\
			}
			$s_t \leftarrow s_{t+1}$, $t \leftarrow t+1$;\\
		}
		$loop \leftarrow loop + 1$
	}
\end{algorithm}

\subsection{DDQN-based Offloading Decision}
In this subsection, we reformulate the first sub-problem, the offloading decision problem, into a Markov decision process (MDP), then propose a DDQN-based offloading scheme to dynamically generate the appropriate strategy to determine $\alpha_t$, $\beta_t$ and tracking mode $k$.

For each time slot $t$, the offloading strategy derived by TAODM only depends on the states consisting of hash similarity, bandwidth, tracking complexity and continuous tracking time of the last time slot $t-1$, which means that the state transition satisfies the Markov property\cite{9963701}\cite{10286340}. Thus, the offloading decision problem can be formulated as an MDP and defined by a tuple $\{S_t, A_t, R_t\}$, which represents the state space, action space and reward space, respectively.

 1) $\emph{State}$: For TAODM, its state $s_t \in S_t$ in time slot $t$ can be represented as $s_t = \{h_t, b_t, c_t, p_t\}$, where $h_t$ is the hash similarity between $f_t$ and the last processed frame $f_{t-1}$, $b_t$ indicates the bandwidth at time slot $t$, $c_t$ stands for tracking complexity that is the number of objects detected by the advanced DNN model in the last offloading frame and $p_t$ is the number of continuous time slots of processing frames with the same tracking mode.

2) $\emph{Action}$: At time slot $t$, the action of TAODM is frame processing mode selection. The action space is formulated as the set of five possible decisions of offloading choices $A_t = \{\emph{Skip, KCF, CSRT, Offload-Full, Offload-ROI}\}$, where \emph{Offload-full} indicates offloading the full frame and \emph{Offload-ROI} indicates offloading ROI blocks. In pursuit of solution efficiency, we merge the ROI decision into TAODM.

3) $\emph{Reward}$: The reward $R$ for TAODM is defined as the weighted sum of actual accuracy $acc_t$ and completion time $l_t^{total}$, which can be expressed as $R = acc_t + \lambda \max((l^{max}-l_t^{total})/l^{max},0)$,
where $acc_t$ is the actual accuracy of frame $f_t$ given by environment, which is used to replace the sum of $acc(m,r)$ and $acc(k)$, and $\lambda > 0$ is the parameter that controls the weight to balance completion time and detection accuracy.

For TAODM's MDP, the deep Q network (DQN) could be applied to train the agent for discrete action space, which leverages the deep neural network to analyze states and actions to optimize the $Q$ value and has a low computation time. In the DQN training process, TAODM first observes the state $s_t \in \mathcal{S}_t$ from the environment and selects the best action $a^{*}_{t} \in \mathcal{A}_t$ for maximizing the action-value function, and can be expressed as $a^{*}_{t}=\argmax_{a_t}{\mathbb{E}[r_t+\gamma\max_{a_{t+1}}Q(s_{t+1}, a_{t+1}; \theta_t)]}$,
where $\theta_t$ is the network parameters matrix and is updated by a back propagation training process, whose loss function is $L(\theta_t) = \mathbb{E}[(y_{t} - Q(s_t, a_t;\theta_t))^2]$,
where $y_t$ is the target value for time slot $t$.
However, the DQN-based algorithm may cause a large deviation in its model due to overestimating the value of Q-target, which indicates the quality of a strategy. To avoid such overestimation, we present a DDQN-based algorithm, which decouples the action selection using DQN and evaluation of Q-target using target network. In the DDQN-based algorithm, the value of Q-target is calculated by $y_{t} = r_{t} + \gamma Q\left(s_{t+1}, \argmax _{a_{t+1}} Q(s_{t+1}, a_{t+1}; \theta_t) ; \theta^{\prime}_t \right)$,
where $\theta_t^{\prime}$ is the target network parameters matrix and is updated periodically. 

An overview of DDQN-based offloading approach is given in Algorithm 1. In each training step $loop$, the specific process is to first initialize hyperparameters, network parameters and replay buffer. Second, TAODM observes the state $s_t \in \mathcal S_t$ from the environment and select the action using $\epsilon^{DDQN}$-greedy method. Then, TAODM executes the action $a_t$ and calculates the reward $R$. Third, a replay buffer $B^d$ is adopted to store the tuple ($s_t$, $a_t$, $R$) for learning Q-network better. Finally, TAODM calculate the Q-value and loss to update network parameters matrix $\theta_t$, but update target network parameters matrix $\theta_t^{\prime}$ every $C$ times.

\subsection{CMAB-based Adaptive Configurations Selection}
\begin{algorithm}[!t]
	\small
	\caption{DDQN-CMAB RL Framework}
	\label{algorithm2}
	\KwIn{Pre-trained MAB models}
	\KwOut{DDQN patameters $\theta$, MAB models}
	Initialize: maximum training steps $LOOP'$ and $loop' = 0$\\
	\For{$loop' \leq LOOP'$}{
		Offloading $f_0$ initialize last detection results; \\
		\For {$t=1$ to $T$} {
			Get new frame $f_t$, bandwidth $b_t \sim N(\rho, \sigma)$, tracking complexity $c_t$, continuous tracking times $p_t$;\\
			Calculate hash similarity $h_t$;\\
			$a_t \leftarrow$ TAODM($h_t, b_t, c_t, p_t$);\\
			
			\If{$a_t = Skip$ }{ 
				Get the last detection results as results of $f_t$;
			}\ElseIf{$a_t = KCF$ or $a_t = CSRT$}{
				TAODM excute tracking algorithm corresponding to $a_t$;\\ 
				Get tracking detection results as results of $f_t$;\\
			}\Else{
				Detected by lightweight DNN model;\\
				\If{$a_t=$ Offload-ROI}{
					extract ROI blocks;\\
				}
				\For {each block $i$} {
					Get information intensity $n_{t,i}$;\\
					$g_{t,i} \leftarrow$ ROIM($n_{t,i}, b_t$);\\
					Offload block $i$ with $g_{t, i}$ to ES;\\				
				}
				Merge detection results of blocks;\\
				Calculate MAB reward $R^{e,c} \ \forall e, c$; \\
				Update gain estimates $Q^{e,c} \ \forall e, c$; \\
				% Update decision counts $N^{e,c} \ \forall e, c$;			
			}
			Perform Algorithm 1 to update $\theta_t$ and $\theta_t^{\prime}$;\\
			$t \leftarrow t+1$;\\
		}
		
		$loop' \leftarrow loop' + 1$;
	}
\end{algorithm}

In this subsection, we reformulate the second sub-problem, adaptive configurations selection, into a contextual MAB problem and employ several multi-armed bandit models to dynamically decide the configurations $g_{t,i} = \{y_{t,i}^{m}, z_{t,i}^{r}\}$, where $y^m_{t,i}$ represents the detection model and $z^r_{t,i}$ represents the offloading resolution for each block $p_{t,i}$.

Based on whether the information density $n_{t,i}$ of block $p_{t,i}$ and bandwidth $b_t$ at time slot $t$ are greater than the average information density $E^i$  and average bandwidth $E^b$, we maintain MABs for four different contexts: i) high information density with high bandwidth ii) low information density with high bandwidth iii) high information density with low bandwidth and iv) low information density with low bandwidth. The motivation behind these four contexts is that if the information density $n_{t,i}$ is lower than the average information density $E^i$, an inferior DNN model would be more likely to achieve similar performance with the advanced one because only few giant objects appear in the block. However, in high information density blocks, multiple small targets tend to appear and utilizing the advanced model will obtain better accuracy. Similarly, if the bandwidth $b_t$ is lower than the average bandwidth $E^b$, the high resolution generates high accuracy but may cause processing failure resulted from an unacceptable transmission delay.

%Let the information density of frame $F_t$ and blocks in time slot $t$ denoted as $I^f_t $ and $I^b_t$. We denote the average information density and bandwidth for time slot $t$ as $E^{I}_t$ and $E^{B}_t$. We use the exponential moving averages method to update the averages with the parameter $\xi \in [0, 1]$ for the most recent information density and bandwidth observation, which defined as follow:%
We use the exponential moving averages method to update the average information density $E^i$ and average bandwidth $E^b$ as follows $E^{i} = \xi_1 n_{t,i} + (1-\xi_1)E^{i}$, $E^{b} = \xi_2 b_t + (1-\xi_2)E^{b}$,
where $\xi_1,\xi_2 \in [0, 1]$ stand for the weight factors for the most recent information density $n_{t,i}$ and bandwidth observation $b_t$, respectively. The above equations give higher weights to the latest information density $n_{t,i}$ and bandwidth ${b_t}$, allowing the model to quickly respond to recent changes.

%Considering that the accuracy of advanced model would likely be higher than low-level one in high information intensity scenario, and the higher offloading resolution would likely be used in high bandwidth scenario, which means that the configurations selection strategy distribution should be different in various contexts. To track the problem, four independent MAB models denoted as $MAB^{HI}_{HB}$, $MAB^{LI}_{HB}$, $MAB^{HI}_{LB}$ and $MAB^{LI}_{LB}$ are maintained to decide the configurations of detection models and offloading resolutions.  
Considering that the configurations selection strategy should be different in various contexts, we maintain four independent MAB models denoted as $MAB^{HI}_{HB}$, $MAB^{LI}_{HB}$, $MAB^{HI}_{LB}$ and $MAB^{LI}_{LB}$ to decide the configurations of detection models and offloading resolutions.  
We use $\mathcal{E} = \{\mathcal{HI, LI}\} \times \{\mathcal{HB, LB}\}$ to denote the context set and use $\mathcal{G} = \mathcal{M} \times \mathcal{R}$ to denote the configurations set. For each context $e \in \mathcal{E}$ and configuration $g \in \mathcal{G}$, we define MAB reward $R^{e,g}$ as $R^{e,g} =R \cdot \mathbbm{1} (e_t = e \wedge g_t = g)$,
where $R$ is the same reward as that in TAODM, $\mathbbm{1}(e_t = e \wedge g_t = g)$ is the contextual indicating function. Thus, each MAB model gets the reward for its decisions allowing independent training. Besides, the reward estimate $Q^{e,g}$ is updated as follows $Q^{e, g} = Q^{e, g} + \varphi \left(R^{e, g}-Q^{e, g}\right), \forall e \in \mathcal{E}, \forall g \in \mathcal{G}$,
where $\varphi$ is the decay parameter. Thereby, each reward estimate $Q^{e,g}$ is updated by the corresponding reward $R^{e,g}$. For all MAB models, a $\epsilon$-greedy based method is used to take decision to train the MAB models as follows

\begin{equation}
	g_{t,i} = \begin{cases}
			\argmax_{g\in \mathcal{G}}Q^{e,g}, &1-\epsilon^{cmab} \\
			\text{random choice}, &\epsilon^{cmab}
		\end{cases} 
	, \quad \forall e \in \mathcal{E},
\end{equation}
where $\epsilon^{cmab}$ is the probability of choosing a random choice. Since we already have precise average estimate of information density and bandwidth after training, we set  $\epsilon^{CMAB} = 0$ to only use exploitation method at in the testing process.

%For this, $\epsilon$-greedy is not a suitable approach as decreasing with time would prevent exploration as time progresses. %Instead, we use an Upper-Confidence-Bound (UCB) exploration strategy that is more suitable as it takes decision counts also into account. Thus, at test time, we take a deterministic decision using the rule
%\begin{equation}
%	a_t = \argmax_{c \in \mathcal{C}}{Q^{e,c}} + \phi \sqrt{\frac{\log t}{N^{e, c}}}, \quad \forall e \in \mathcal{C},
%\end{equation}
%where $\phi$ is the exploration factor and $t$ is the time slot count.

\subsection{Real-time Tracking-assisted Video Analysis Framework}
\begin{figure}[!t]
	\centering
	\includegraphics[width=0.99\linewidth]{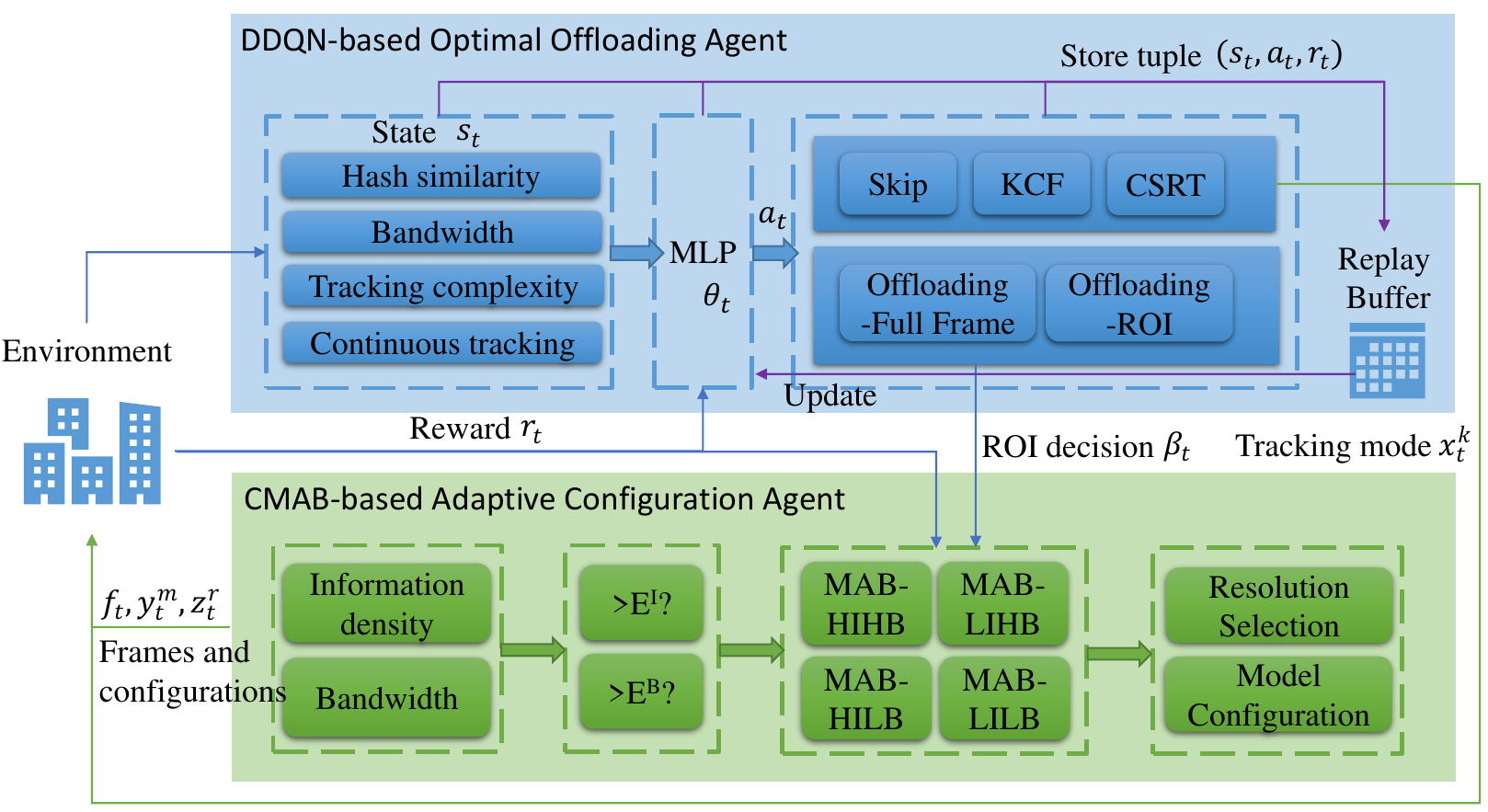}\\
	\caption{The proposed DCRL framework.}
	\label{fig_framework}
\end{figure}

Since TAODM and ROIM are highly coupled and only when TAODM decides to offload the frame to ES, ROIM will be used to decide the configurations for each block. Based on these insights, we propose a double-layer DDQN-CMAB reinforcement learning (DCRL) framework to jointly train DDQN-based offloading agent and CMAB-based configurations selection agent.

As shown in Fig. \ref{fig_framework}, the proposed DCRL framework consists of  two layers: i) in the upper layer, we assign TAODM as the first controller, which observes the hash similarity $h_t$, bandwidth $b_t$, tracking complexity $c_t$ and continuous tracking time $p_t$ from the environment, then makes the offloading decision. If TAODM decides to process the frame locally, the frame results will be obtained with a tracking mode $k$. Otherwise, the frame will be sent to the lower layer; ii) in the lower layer, ROIM observes the information density $n_{t,i}$ and bandwidth $b_t$, and determines the configuration $g_{t,i}$ of detection models $m$ and offloading resolution $r$ for each block.

In each training step of DCRL, we first offload the entire frame of $f_0$ to initialize the last detection results for tracking in the future. Then at each time slot $t$, TAODM determines $a_t$ by DDQN. If the $Skip$ mode is adopted, directly use the last detection results as the detection results of $f_t$. If the KCF or CSRT mode is adopted, TAODM executes the tracking algorithm corresponding to $a_t$ to get detection results. Otherwise, the frame would be processed by ROIM, where CMAB is performed to obtain the configuration $g_{t,i}$ of each block and then offload all blocks to the ES. MAB reward $R^{e,g}$ is calculated to update estimates $Q^{e,g}$ when receiving the results of all blocks. Meanwhile, % decision counts $N^{e,c}$ are updated. Finally, 
TAODM receives detection results, and the rewards of DDQN are calculated for updating its DNN parameters matrix $\theta_t$ and target Q-network parameters matrix $\theta^{\prime}_t$. The detailed steps of the DCRL training framework are summarized in Algorithm \ref{algorithm2}.

\section{Simulation Results}\label{simulation_results}
\begin{figure*}[!ht]
	\centering
	\addtocounter{figure}{-1}
	\label{fig:ThreeFig}
	\subfigure{
		\begin{minipage}[t]{0.23\linewidth}
			\centering
			\includegraphics[width=1.5in]{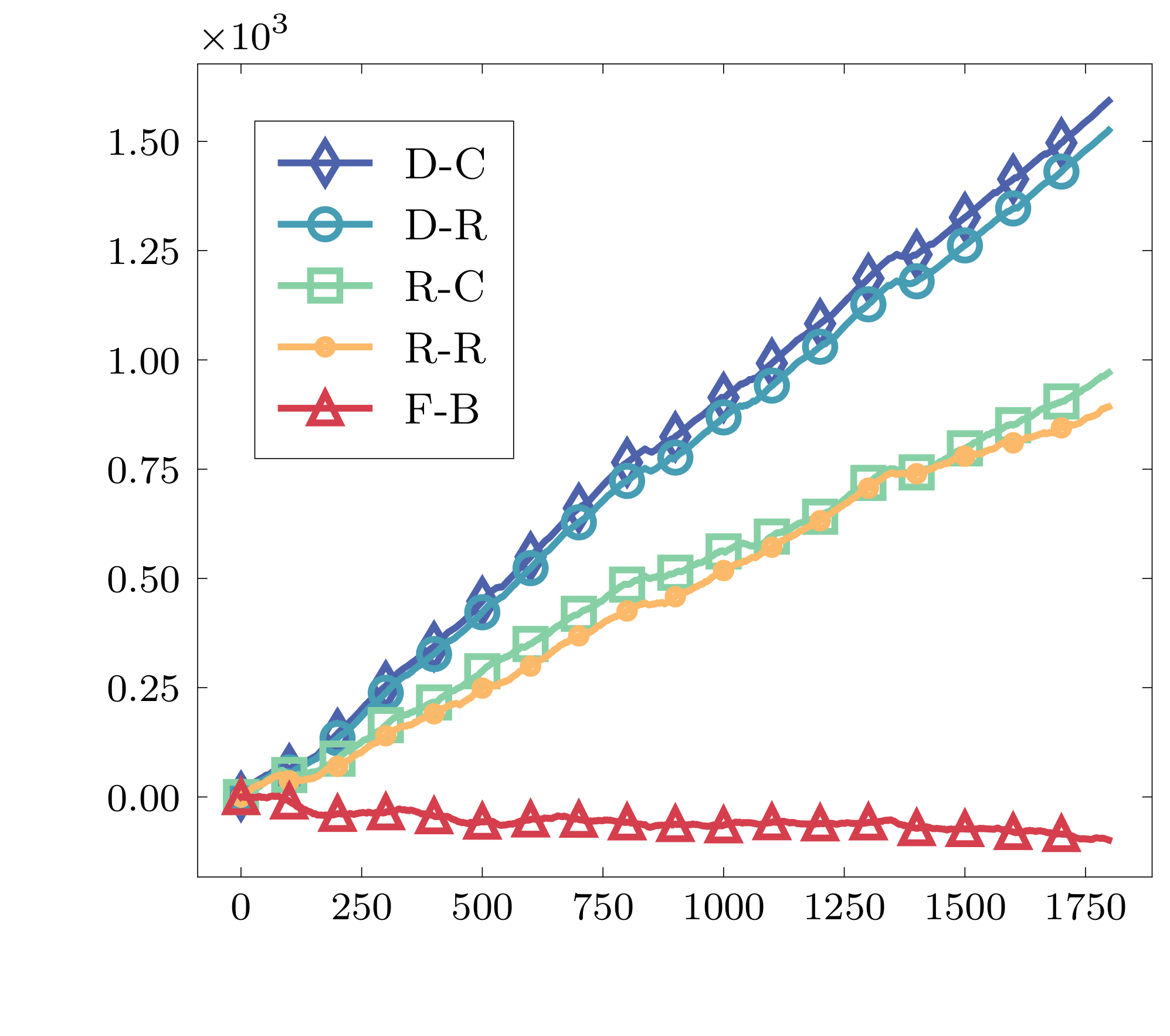}\\
			\caption{Performance comparison on cumulative rewards.}\label{cum_rewards}
		\end{minipage}%
	}%
	\hspace{1mm}
	\subfigure{
		\begin{minipage}[t]{0.23\linewidth}
			\centering
			\includegraphics[width=1.5in]{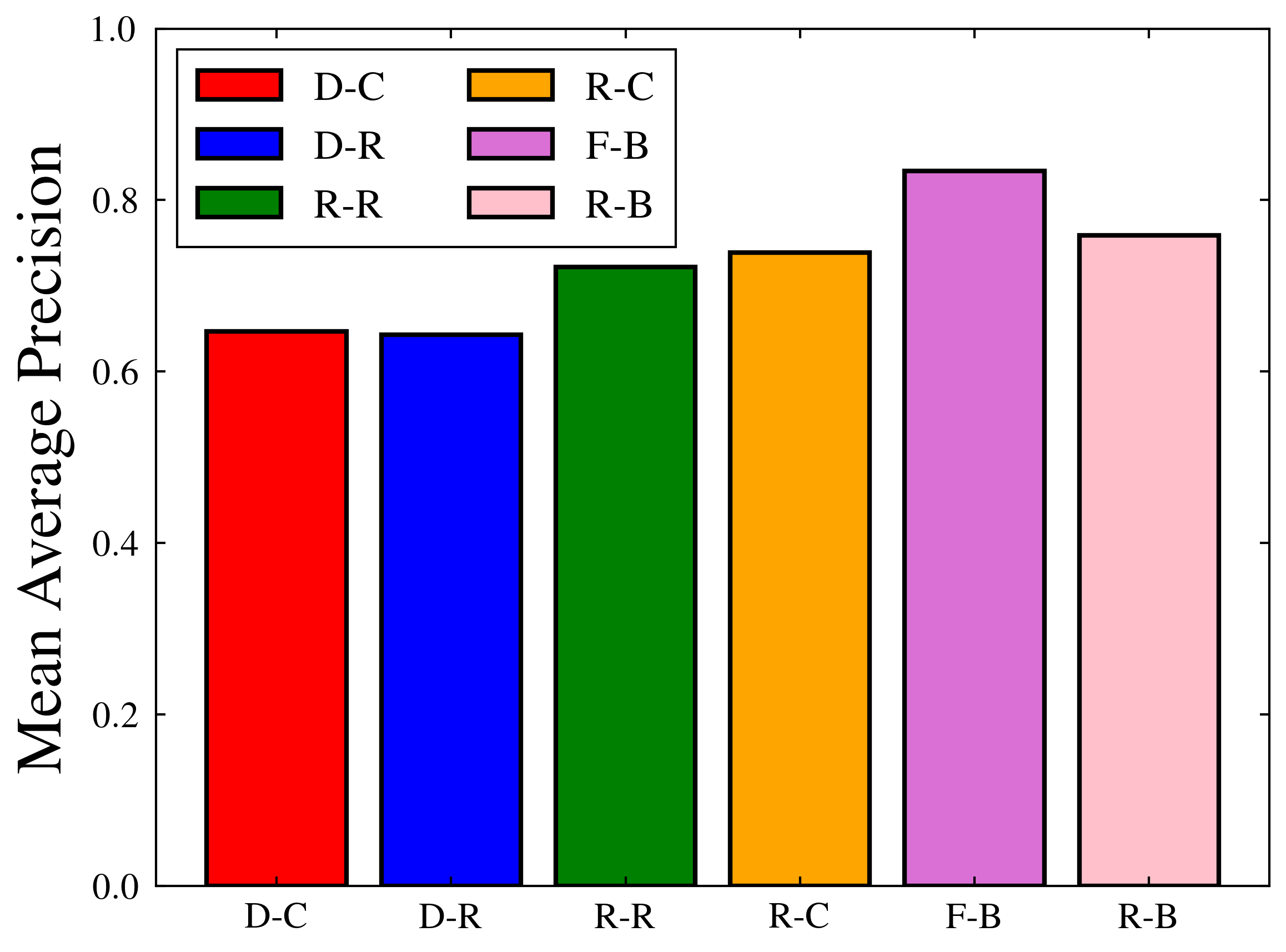}\\
			\caption{Performance comparison on mAP.}\label{accuracy}
		\end{minipage}%
	}%
	\hspace{1mm}
	\subfigure{
		\begin{minipage}[t]{0.23\linewidth}
			\centering
			\includegraphics[width=1.5in]{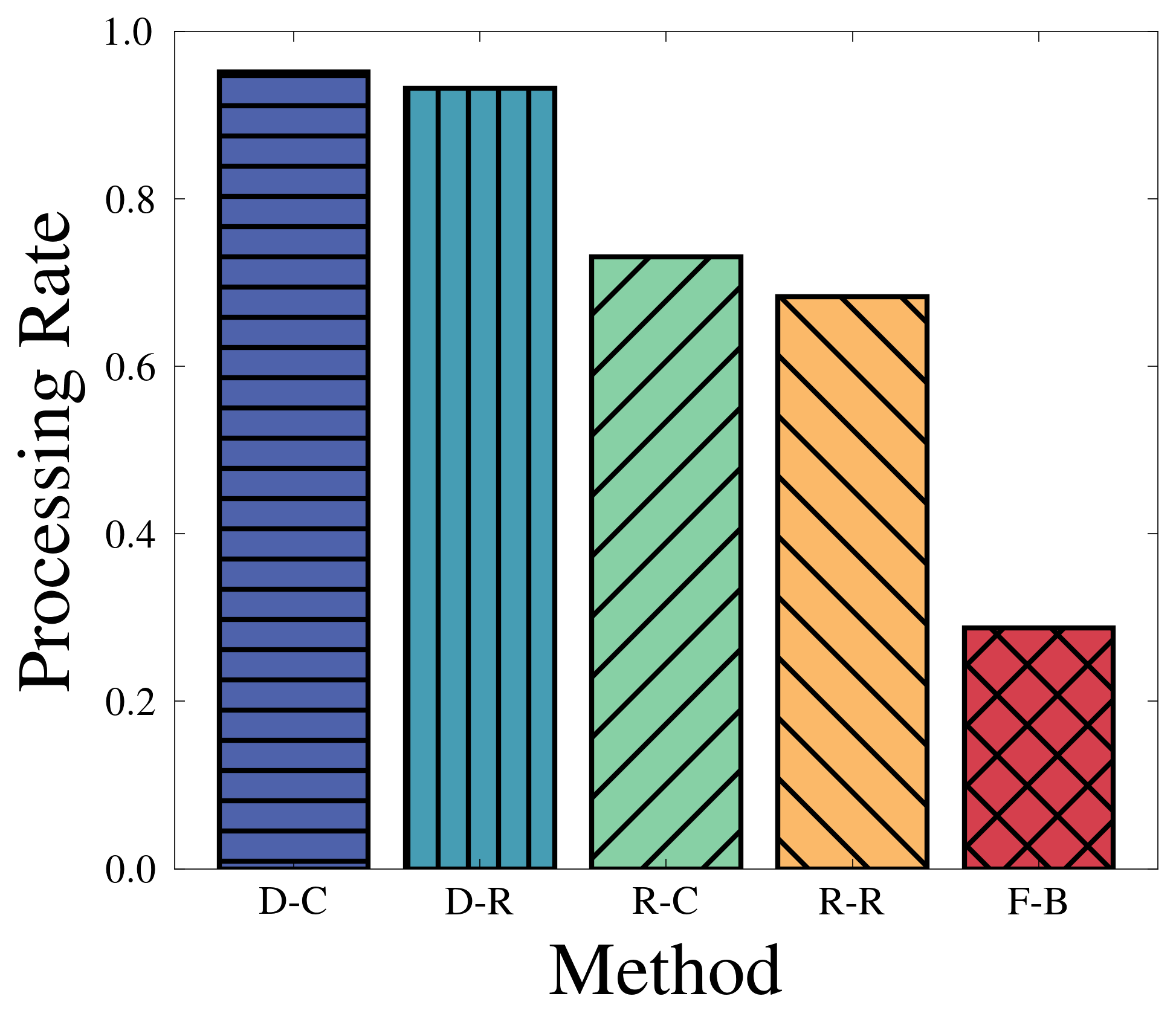}\\
			\caption{Performance comparison on processing rate.}\label{processing_rate}
		\end{minipage}
	}%
	\centering
	\hspace{1mm}
	\subfigure{
		\begin{minipage}[t]{0.23\linewidth}
			\centering
			\includegraphics[width=1.5in]{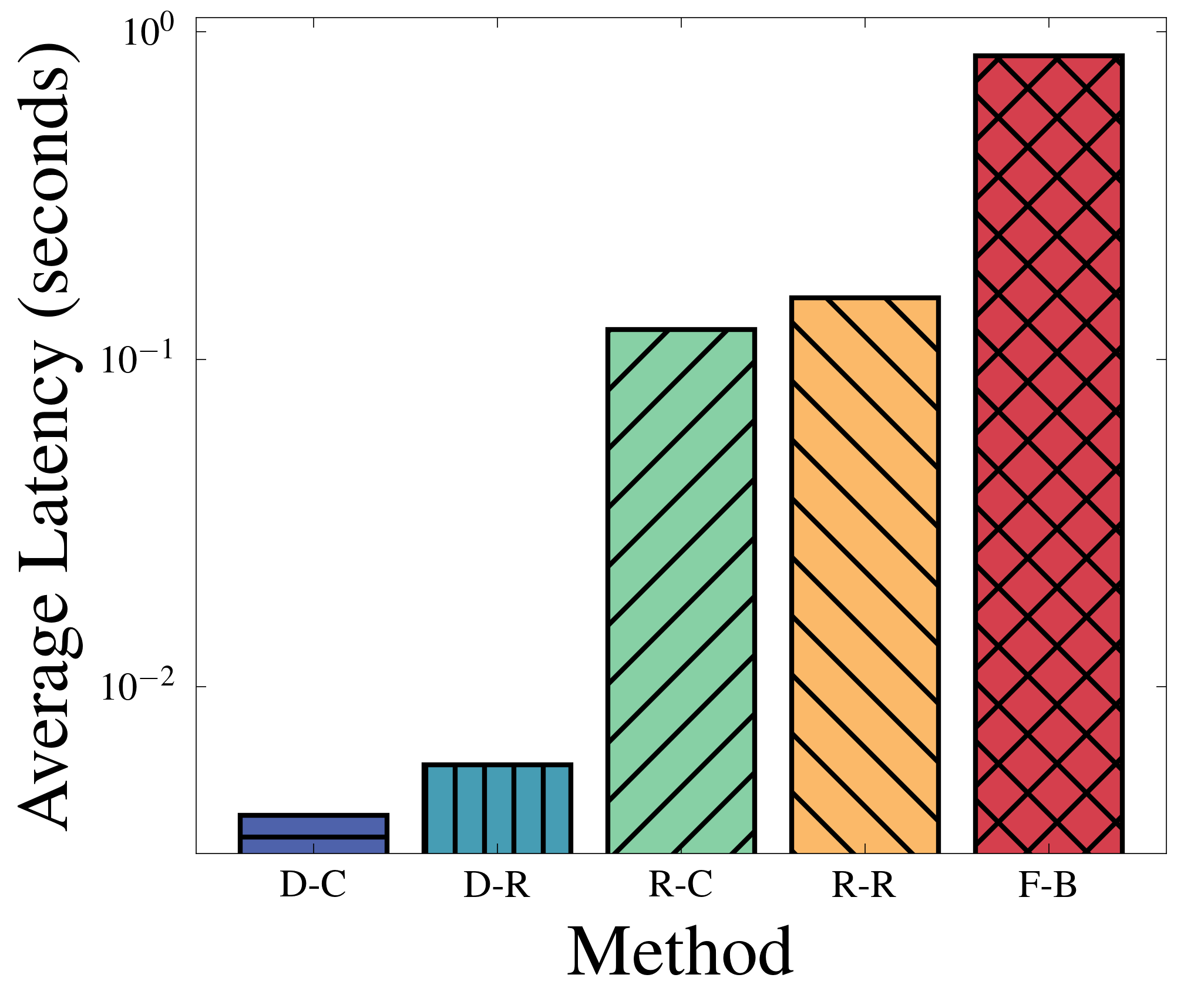}\\
			\caption{Performance comparison on average latency.}\label{latency}
		\end{minipage}
	}%
	\centering
	%\caption{}
\end{figure*}

%\begin{table}[!t]
%	%\footnotesize
%	\caption{PARAMETER SETTINGS}
%	\renewcommand{\arraystretch}{1.2}
%	\label{table1}
%	\centering
%	\footnotesize
%	\begin{tabular}{c|c|c|c|c|c}
%		\hline
%		\bfseries{Param.} & \bfseries{Value} & \bfseries{Param.} & \bfseries{Value} &\bfseries{Param.} & \bfseries{Value}\\
%		\hline
%		$\varphi$  & 0.9 & $\rho$ & 10 & $\eta$ & 1.0\\
%		$\omega$  & 0.9 & $\sigma$ & 5 &$f^{edge}$& 2.0\\
%		$\tau$         &     & $\xi_1$    & $0.9$ & $f^{device}$& 2.0 \\
%		$\lambda$       & $0.3$                    & $\xi_2$    & $0.9$ & $\varepsilon^{ddqn}$ & 0.3\\
%		$\eta$       & $1.0$                  & $\gamma$     & $0.9$ &$\varepsilon^{cmab}$ & 0.3\\
%		\hline
%	\end{tabular}
%\end{table}

%$[5.6e-4, 6.25e-4, 6.5e-4]$ MB/Pixel
In this section, simulations are conducted to evaluate the performance of the proposed DCRL on an open dateset, i.e., Garden2 of Multi-camera Pedestrian Video Dataset\cite{crossdataset}.To ensure fairness, 80\% of each video is used for training and the remaining for testing. Furthermore, let $\tau \in \{5.6e^{-4}, 6.25e^{-4}, 6.5e^{-4}\}$ MB/Pixel, $\varepsilon^{ddqn}=0.3$, $\varepsilon^{cmab}=0.3$, $\rho = 10$, $\sigma=5$, and $f^{edge}:f^{device}=2:1$.

%Similar settings have also been employed in the literature \cite{3}. 
%Note that some parameters may be varied for different evaluation purposes. 
For comparison purposes, except the proposed DCRL, the following schemes are also simulated as benchmarks.
	%\item \textit{Sample-5-Best}: Sample the frame fail counts equals t.
	\textit{RAND-RAND (R-R)}: TAODM and ROIM randomly determine offloading and configurations of detection model and offloading resolution.
	\textit{RAND-CMAB (R-C)}: TAODM randomly determines offloading, while ROIM determines configurations of detection model and offloading resolution using CMAB-based adaptive configurations selection algorithm.
	\textit{DDQN-RAND (D-R)}: ROIM randomly determines configurations of detection model and offloading resolution, while TAODM determines offloading using DDQN-based offloading algorithm.
	\textit{FULL-HIGH (F-B)}: TAODM only adopts \emph{Offload-full} decision and ROIM adopts the highest offloading resolution with the most accurate model configuration.
	%\item \textit{ROI-HIGH (R-B)}: TAODM only adopts \emph{Offload-ROI} decision and ROIM adopts the highest offloading resolution with the most accurate model configuration.
	%\item \textit{Yolov5l-CMAB}: Offloading frames to edge server and only inference by Yolov5m model with CMAB-based configuration selection.
	%\item \textit{Yolov5l-Best}: TAODM offloading all frames to edge server and selects the highest resolution and the most accurate DNN model for all frames to maximize accuracy.
	%\item \textit{DDQN-CMAB}: TAODM offloading all frames to edge server and selects the highest resolution and the most accurate DNN model for all frames to maximize accuracy.
%Fig. 3 examines the convergence of the proposed HDRL training framework integrating both the DQN-based optimal routing and adaptive key frames selection approaches. demonstrate the superiority of our approach in terms of cumulative rewards.

Fig. \ref{cum_rewards} shows the cumulative rewards with time, it can be observed the proposed \emph{DDQN-CMAB (D-C)} outperforms the other benchmarks, as it filters the repeated spatial-temporal semantic information to ensure the results from ES return in time while obtaining high accuracy. In addition, \emph{F-B} obtains the worst performance because offloading full frames generates a large transmission delay under fluctuating network conditions.

Fig. \ref{accuracy} illustrates the mean average precision (mAP) with different methods, we can see \emph{D-C} obtains the highest accuracy except for \emph{F-B}. Fig. \ref{processing_rate} compares the processing rate of different methods, and it can be seen that \emph{D-C} achieves the best processing rate compared with others. Specifically, DCRL improves the processing rate by up to 66.3\% compared to \emph{F-B}. 
Compare Fig. \ref{accuracy} to Fig. \ref{processing_rate}, it can be observed that i) \emph{F-H} realizes the best mAP performance while the lowest processing rate for large transmission delays, ii) \emph{R-H} achieves higher mAP with higher processing rate since the repeated spatial-temporal semantic information is filtered, where the effectiveness of ROI is fully illustrated.

Fig. \ref{latency} compares the average latency of different methods, and \emph{D-C} achieves the lowest latency compared with others. Exploring the reasons, we find that TAODM adopts \emph{offload-ROI} to minimize the adverse effects of fluctuating network conditions.
%Fig. \ref{accuracy} and \ref{latency} illustrates the mean average precision and the processing rate with different methods. Compare Fig. \ref{accuracy} to Fig. \ref{latency}, it can be observed that i) in Fig. \ref{cum_rewards}, the proposed approach outperforms all benchmarks in terms of cumulative rewards; ii) in Fig. \ref{accuracy}, DCRL achieves higher mean average accuracy compared with others.
%Specifically, DCRL improves the utility by up to 70.0\% compared to the Local algorithm, and by up to 22.3\% compared to the best baseline algorithm (i.e., LinUCB algorithm). Combining the advantages of AKMA and NRSA, VQODAP performs the best regardless of load. 
When network conditions are stressful, ROIM can effectively reduce data volume to ensure that critical information is offloaded to the ES for accurate detection. Besides, dramatic network fluctuations are often temporary, and a local tracking algorithm is used to tide over this difficult period to improve accuracy when network conditions are poor.

\section{Conclusion}\label{conclusion}
In this paper, a DCRL framework integrating a DDQN-based optimal offloading decision approach and a CMAB-based adaptive configuration selection approach is proposed to address the challenge of balancing frame processing rate and accuracy performance of edge-based real-time video analysis systems in intelligent visual devices. By decomposing the optimization problem of video analysis into two subproblems and generating an optimal strategy for offloading mode and the configurations of detection model and resolution selection, our proposed framework can effectively solve the overall optimization problem. Experimental results show that our approach outperforms counterparts in terms of ensuring a high process rate with high detection accuracy. 
\begin{spacing}{0.9}
	\bibliographystyle{IEEEtran}
	\bibliography{IEEEabrv,Ref_VQODAP}
\end{spacing}

\end{document}